\documentclass[conference]{ASYU}
\IEEEoverridecommandlockouts

\usepackage[utf8]{inputenc} 
\usepackage[T1]{fontenc}

\usepackage{comment}
\ifCLASSINFOpdf
  \usepackage[pdftex]{graphicx}
\else
\fi
\usepackage[cmex10]{amsmath}
\usepackage{multirow}
\usepackage{array}
\usepackage[lofdepth,lotdepth]{subfig}

\DeclareMathOperator*{\argmin}{arg\,min}
\newcommand\blfootnote[1]{%
  \begingroup
  \renewcommand\thefootnote{}\footnote{#1}%
  \addtocounter{footnote}{-1}%
  \endgroup
}

\begin{document}


%
\title{Detecting Turkish Synonyms  Used in Different Time Periods}

\author{\IEEEauthorblockN{Umur Togay Yazar$^{1}$, Mucahid Kutlu$^{2}$} 
\IEEEauthorblockA{$^{1}$Computer Engineering Department, TOBB University of Economics and Technology, Ankara, Türkiye\\
u.yazar@etu.edu.tr \\
$^{2}$Computer Science and Engineering Department, Qatar University, Doha, Qatar\\
mucahidkutlu@qu.edu.qa
}
}

%

\maketitle


\begin{abstract}
Dynamic structure of languages  poses significant challenges in applying  natural language processing models on historical texts, causing  decreased performance in various downstream tasks.  Turkish is a prominent example of rapid linguistic transformation due to the language reform in the 20th century.  In this paper, we propose two methods for detecting synonyms used in different time periods, focusing on Turkish.  In our first method, we use Orthogonal Procrustes method to align the embedding spaces created  using documents written in the corresponding time periods. In our second method, we extend the first one by incorporating Spearman's correlation between frequencies of words throughout the years. In our experiments, we show that our proposed methods outperform the baseline method. Furthermore, we observe that the efficacy of our methods remains consistent when the target time period shifts from the 1960s to the 1980s. However, their performance slightly decreases for subsequent time periods. 

\end{abstract}



%
\IEEEpeerreviewmaketitle

\IEEEpubidadjcol

\section{Introduction}
\blfootnote{979-8-3503-7943-3/24/\$31.00 ©2024 IEEE} 
Language 
is an essential phenomenon for living creatures, providing a communication channel to convey meaning. 
One of the most prominent characteristics of languages is their dynamic structure, i.e., the components that form a language change over time \cite{aitc}. These changes are shaped by the society in which the language exists, reflecting socio-cultural properties. 

Turkish is a good example in which one can easily observe the language evolution resulting from sociocultural and sociopolitical actions in a relatively short time period due to the Turkish language reform 
\cite{catastrophic}. Many  alterations have been proposed as a reflection of modernization efforts; leading to various types of linguistic changes in Turkish. 
The first step  was taken in 1928 by switching to the Latin alphabet. The second significant attempt in Turkish includes the efforts to purify the language by substituting Persian and Arabic origin words, which were numerous during the Ottoman era, with words from Turkish origins. 
 
The changes in a language make it harder to analyze and understand historical documents. For instance, due to the extensive \textit{neologism} \cite{lex,progress}, i.e., introducing a new term for a concept in the language, within the Turkish language reform, many people, especially younger generations, are not able to understand the texts written in the Ottoman era and the first decades of Türkiye. Furthermore, dictionaries generally do not contain the entire lexicon, making them inherently incomplete.

For example, the words \textit{müteşir, bab} and \textit{vecahi} does not exist in the current version of the dictionary of Turkish Language Association (TDK)\footnote{https://sozluk.gov.tr/}. 

Moreover,  changes in a language will also reduce the application of natural language processing (NLP) tools for historical documents.
The performance of large language models (LLMs) significantly decreases when they encounter words or phrases from a different time period than that of its training data, i.e., temporal misalignment \cite{temporal_mis}. In addition, this situation can decrease the accuracy and reliability of the models in the downstream tasks such as  named entity recognition \cite{temporal_mis,entity} and question answering \cite{test_data}. Recently, Zheng et al. \cite{test_data} report that a dramatic performance decrease in machine translation can be observed when neologism words are included in the text.  Therefore, we need models that help us to detect the meaning of a given word in a historical document and analyze these documents in an effective way.  

In this paper, we work on detecting semantically similar words for a given word used in a different time periods for Turkish. We propose two methods, namely \textit{Orthogonal Procrustes Alignment} (OP) and 
\textit{Orthogonal Procrustes with Spearman rank correlation} (OP+SC). OP  is based on finding the transformation matrix between the embedding spaces of two different time periods. 
OP+SC is a novel improvement to the OP that involves a selection mechanism on top of the results returned by OP based on Spearman's rank correlation scores of word pairs.

In our experiments, we use Turkronicles \cite{turkronicles}  which is a Turkish diachronic corpus consisting of documents from 1920 to 2020s.  
We evaluate the performance of the proposed methods using  Continuous Bag of Words (CBOW) and Singular Value Decomposition (SVD) embeddings.  Furthermore, we assess the robustness of these methods against varying time distances between the base and target periods. 
In our experiments we show that our proposed methods outperform the baseline method proposed by Zhang et al. \cite{omni} to detect analogical counterparts. 
In addition, incorporating Spearman's correlation coefficient  increases the performance of the OP model  with both CBOW and SVD embeddings in most of the cases. 
Lastly, we find that the performance of our methods remains similar when the target time period is changed from 1960s to 1980s. However, their performance tends to gradually decrease over the subsequent time periods. 

The contributions of our work as follows.
\begin{itemize}
    \item Although many computational studies examining various subjects of linguistics in English exist, to the best of our knowledge, our study is the  first study  that focuses on neologism in Turkish with natural language processing methods. 
    \item We propose two approaches to detect synonyms  used in different time periods. In our experiments, we show that our proposed methods outperform the baseline method.
    \item We share our code to maintain the reproducibility of our results\footnote{https://github.com/togayyazar/Lingan}.  
\end{itemize}

The rest of the paper is organized as follows. In Section \ref{sec:related}, we present related works. In Section \ref{sec:problem_definition}, we formally define the problem and introduce the mathematical notations used in the paper. In Section \ref{sec:methodology}, we explain our proposed methods. 
We present the results of our experiments in  Section \ref{sec:exp_res}. In Section \ref{sec:limitations}, we discuss the limitations of the proposed methods and  conclude in Section \ref{sec:conclusion}.

\section{Related Works}\label{sec:related}
Neologism refers to creating new words or concepts, often driven by internal, external, or semantic mechanisms \cite{neo_def}. In linguistics, neologism is an important concept because it sheds light on the mechanisms by which languages adapt to new cultural and technological changes \cite{neo_1}.  Additionally, understanding neologisms helps linguists track language change and evolution in the context of interaction between languages \cite{bybee}. In the literature, there exist studies that approach neologism from a computational perspective. Pechenick et al. \cite{pechenick} and Michel et al. \cite{michel} perform a diachronic analysis on English, observing changes in the lexicon over the years by comparing English decade-by-decade using frequency-based methods. Del Tredici and  Fernández \cite{tredici} explore the graph structure that affects the emergence of new words and assesses how lexical innovations arise and spread, using social network analysis techniques. 

With the implementation of distributional semantics methods, dense vector representations \cite{mikolov,pennington} have been effectively used in NLP studies, enabling the testing of various linguistic hypotheses \cite{xu,neology}. One of the main areas where word embeddings are extensively utilized is the diachronic analysis, particularly the semantic change \cite{kutuzov}, i.e., tracking the change in the meaning of a word through time. In semantic change, typically, a diachronic corpus is split into discrete time intervals and a vector space for each interval is obtained using different types of word embedding algorithms \cite{kulkarni,hamilton_a,hamilton_b,rudolph}. After the embeddings are generated, the embedding spaces are aligned to perform vector comparisons across different time periods. There exist various methods to align embedding spaces such as applying linear transformations \cite{kulkarni},  finding an orthogonal transformation matrix by Orthogonal Procrustes alignment \cite{hamilton_a,hamilton_b}, and  aligning the word vector of a query word by involving the somehow related words in the target time period \cite{omni}. 
As another approach, a number of researchers explored jointly training the word embeddings by considering the alignment during the training phase \cite{dyn1,dyn2}. In our work, we use the Orthogonal Procrustes Alignment as in \cite{hamilton_a}, but we focus on a different form of linguistic change. In addition, we extend the method they use by incorporating a selection logic based on the Spearman's rank correlation.

Several researchers approached neologism using word embeddings. Zhang et al. \cite{omni} find the counterpart of a query word by performing local and global correspondence matching. In the global correspondence approach, they obtain a transformation matrix by minimizing the Euclidean distance between the temporal representations of the seed words. In the local correspondence approach, they construct a semantic graph representation of the query word to find its counterpart. However, Zhang et al. \cite{hierarchy} observe a performance drop on different types of words when applying the same global transformation to all words. They create a transformation matrix for each word which outperforms the global transformation approach.  These studies aim to find the analogical counterparts,  e.g., retrieving \textit{Walkman} from 1987-1991 by querying the word \textit{iPod} from 2002-2007 \cite{omni}. This is different from our case because finding analogical counterparts requires that 
the matched words belong to similar or related semantic fields, given that the exact concept of the query words may not have emerged in the previous time periods. In contrast, we focus on the words borrowed mainly from mostly Arabic and Persian languages and replaced with modern Turkish words  due to the Turkish Language Reform. Thus, our goal is to find  words that  exactly correspond to the same concept or meaning, such as matching \textit{belge}-\textit{vesika},  rather than word pairs affected by technological advancements. Another difference is that all these works use English corpora. To the best of our knowledge, our study is the first study to examine neologism in Turkish from a computational perspective.

\section{Problem Definition}\label{sec:problem_definition}

Our goal is to identify synonyms used in documents written in two different periods, enabling us to understand and analyze historical documents. 
We formulate our problem as a ranking problem: for a given word  $w$  used in documents written in the \textit{base} time period $T_b$, rank the words used in documents written in the \textit{target} time period $T_t$ based on their semantic similarity with respect to the word $w$.

It is worth noting that the problem extends beyond simply finding similar words in the vector spaces. Firstly, the meaning of a word may change over time. Thus, even if a word exist in both time periods, the meanings might be different in each period. Furthermore, 
even if a word does not undergo semantic change, its representations from different time periods are not directly comparable due to  the transformations applied by the word-embedding algorithms, e.g. the randomized initialization in  Continuous bag-of-words (CBOW) and the non-unique nature of  Singular Value Decomposition (SVD) \cite{kulkarni,hamilton_a}.

\section{Proposed Methods}\label{sec:methodology}
In this section, we explain our proposed methods  to identify the synonyms in two different time periods. 


\subsection{Orthogonal Procrustes  Alignment (OP)}
Intuitively, when a new word is introduced to a language to replace an existing one, it should at least bear as much semantic information as the existing one. Therefore, we should be able to use the new word within the same context of the replaced word  
without altering the meaning. 
Motivated by this intuition, in this method, we first find word embedding vectors using the documents written in the time periods of $T_b$ and $T_t$, separately. We use $\mathbf{W}_{T_b}$, $\mathbf{W}_{T_t}$ notations for these word embedding spaces. Subsequently, we find the most similar vectors in $\mathbf{W}_{T_t}$ to the vector of $w$ in $\mathbf{W}_{T_b}$. However, a vector from $\mathbf{W}_{T_b}$ cannot be directly compared with a vector from $\mathbf{W}_{T_t}$ due to arbitrarily transformed axes \cite{kulkarni,hamilton_a}. Thus, the two vector spaces must be aligned with each other to ensure proper comparison between embeddings. Furthermore, a word in $T_b$ may not exist in time period $T_t$. Thus,  if a word is not present, its vector representation in $T_t$ should be still constructed in such a way that conveys a similar meaning to the base period $T_b$ to perform vector operations. 

Based on these needs, we use Orthogonal
Procrustes \cite{schonemann} to align the embedding spaces $\mathbf{W}_{T_b}$ to $\mathbf{W}_{T_t}$. 
In particular, OP tries to minimize the difference between the 
vectors of $W_{T_b}$ and $W_{T_t}$ by finding an orthogonal transformation matrix \textit{Q}. The objective function is formulated as follows. 

\begin{equation}\label{eq:op}
\argmin_{\mathbf{Q^TQ=I}} \| \mathbf{E}_{T_b}\mathbf{Q} - \mathbf{E}_{T_t} \|^2_F
\end{equation}

\noindent
where $\mathbf{Q} \in R^{d\times d}$ is the orthogonal transformation matrix that bridges  $T_b$ and $T_t$. $\mathbf{E}_{T_b}$ and $\mathbf{E}_{T_t}$ are the intersection embeddings, created by taking the intersection of the vocabularies of the time periods and aligning the corresponding vectors by index.  The optimal solution can be found through the application of SVD on the matrix $\mathbf{K=W}_{T_b}\mathbf{W}_{T_t}^\textit{T}$. 

After finding $\mathbf{Q}$, we align the vector $\mathbf{w} \in \mathbf{W}_{T_b} $ of the query word $w$ performing the transformation $\mathbf{Q}\mathbf{w}$. The expectation is that the aligned vector of $w$ should be close to the vector of its counterpart $\mathbf{w'} \in \mathbf{W}_{T_t}$, and thus, falls into the same semantic field of $w'$ in time period $T_t$. Finally, we rank the words and perform a k-nearest neighbors search to detect the words semantically similar to the $w$.

\subsection{Orthogonal Procrustes Alignment with Spearman's Rank Correlation (OP+SC)}
\begin{figure}
    \centering
    \includegraphics[width=1\linewidth]{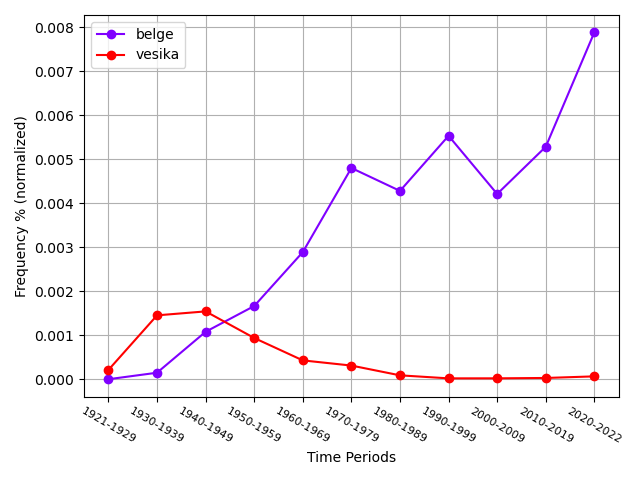}
    \caption{Relative frequency  of words \textit{belge} and \textit{vesika} between 1920 and 2022 in ten-year periods in Turkronicles dataset.}
    \label{fig:belge_vesika}
\end{figure}

Since the second half of the 1920s, many new words of Turkish origin have been included in the language, replacing  words borrowed from other languages. Consequently, it is reasonable to expect a decline in the prevalence of borrowed words over time, while the frequency of newly introduced words increases. As a showcase example, \textit{belge} is a new Turkish word introduced as a substitute for the Arabic-originated word \textit{vesika}. The change in their frequencies from the 1920s to the 2020s in the Turkronicles \cite{turkronicles} dataset is shown in Figure \ref{fig:belge_vesika}.

We incorporate this observation to improve the performance of OP. First, we use  OP  to find $k$-nearest neighbors set $K$ in $T_t$ for the query word $w$. Next, a frequency time series is constructed for each candidate pair ($w$, $c$), where $c \in K$. Subsequently, we re-rank  the pairs according to Spearman's correlation coefficient in decreasing order. This method increases the rank of the true counterpart of $w$ in the $k$-nearest neighbor set. 

\section{Experiments}\label{sec:exp_res}

\subsection{Experimental Setup}

\noindent
\textbf{Dataset.}
In our experiments, we utilized the Turkronicles dataset \cite{turkronicles} which is  a  diachronic corpus consisting of the records of the Grand National Assembly of Türkiye (TBMM) and the Official Gazette of Türkiye (known as \textit{Resmi Gazete}) from 1920s to 2020s. The corpus contains 45K documents  and 850M tokens in total, and  each document has a time-stamp, allowing us to observe changes in Turkish over time. 
Furthermore, as the documents in the corpus are issued by the state, the language used is  formal and the documents reflect the linguistic evolution shaped by governmental policies. 

In order to evaluate the performance of our proposed methods, we manually created a dataset.  As it was challenging to identify the list of the new words with their old counterparts, we conducted the following strategy. Firstly,  we determined the base period as 1930 and the target period as 1980. Subsequently, we constructed a unigram distribution for the words within these selected time spans (i.e., 1930-1939 and 1980-1989). Next, using the unigram distributions, we calculated the Jensen Shannon Divergence (JSD) between the distributions as in  Equation  (\ref{eq:jsd}) to detect the vocabulary change. 

\begin{equation}\label{eq:jsd}
JSD(P_T||P_B) = \frac{1}{2} KL(P_T||A) + \frac{1}{2} KL(P_B||A)   
\end{equation}

\begin{equation}
 A=\frac{1}{2}\left(P_T+P_B\right)
\end{equation}

\noindent
where $A$ is the average distribution, $P_T$ and $P_B$ are respectively the unigram distribution of the target period and base period, and $KL\left(\cdot||A\right)$ is the \textit{Kullback–Leibler} divergence.
In this calculation, we selected the top 5,000 words that contributed the most to the divergence score. We then categorized these words based on their relative frequencies in the respective time periods. This process enables us to identify words whose frequency changed dramatically between these periods, indicating potentially replaced or newly added words.

Subsequently, using the TDK dictionary, we manually created pairs of words ($w$, $w'$), where $w$ is a word in the target period and $w'$ is its old counterpart in the base time period among these 5,000 words. Eventually, we created a dataset with 221 pairs.

\noindent 
\textbf{Implementation.} In our experiments, we  explore the impact of different parameters of our methods. In particular, the embeddings used in this study are diachronic, i.e., the vectors are obtained for each 10-year time period from 1921 to 2022. We use SVD  and CBOW embeddings.  
SVD and CBOW embeddings are both $d=300$ dimensional. The smoothed unigram distribution with $\alpha=0.75$ was incorporated into the training phase in both embeddings to mitigate the negative effects of low-frequency words \cite{levy}. The context window size is 2, used with a symmetric kernel both in SVD and CBOW. For CBOW embeddings, the number of negative samples is set to 5, and the downsampling rate is $1\times 10^{-5}$. 

\noindent
\textbf{Baseline.} We compare our proposed method with the  linear transformation (LT) approach of \cite{omni} as our baseline method. In finding the analogical counterparts with the LT, the core part is creating a linear transformation matrix between the embedding spaces of time periods being aligned. The transformation matrix is obtained as follows. 

\begin{equation}\label{eq:lt}
\argmin_{\mathbf{M}} \sum_{x \in S} \| \mathbf{M}\mathbf{x}_b - \mathbf{x}_t \|^2 + \alpha\|\mathbf{M}\|^2
\end{equation}

\noindent
where  $\mathbf{M}$ is the linear transformation matrix, $S$ is the set that contains seed pairs, $\mathbf{x}_b$ and $\mathbf{x}_t$ is the vector representations of $x \in S$ in base and target period, respectively.  In \cite{omni}, seed pairs are defined as words that are 
common in both the base and target time periods and in the first 5\% portion of words by frequency. In our experiments, we select the seed pairs from the most frequent 1,000 words. Finally, we train a Ridge Regression model with $\alpha = 0.2$ to find M.

\noindent
\textbf{Evaluation Metrics.} 
We formulate our problem as a ranking problem in which there is only one correct word for each query word. Therefore, we adopt  Recall and Mean Reciprocal Rank (MRR) metrics from information retrieval field to quantify the performance of the methods. MRR is an effective method when there is only a single correct answer. It can be calculated as follows. 

\begin{equation}
    \text{MRR} = \frac{1}{|N|} \sum_{i=1}^{|N|} \frac{1}{\text{rank}_i}
\end{equation}

\noindent
where  $|N|$ is the number of queries and  $\text{rank}_i$ is the rank position of the first correct word  for the $i$-th query. Recall is used to measure how well the system retrieves the correct items within the top K results and it is formulated as follows. 

\begin{equation}
    \text{Recall@k} = \frac{TP_k}{TP_k + FN_k}
\end{equation}

\noindent 
where $TP_k$ is the number of true positives within the top $k$ results and $FN_k$ is the number of false negatives within the top $k$ results. In our experiments, we set K=1, 10, and 100. For our OP+SC method, we  get the nearest 5, 15, and 150 neighbors for Recall@1, Recall@10, Recall@100 metrics, respectively.

\subsection{Results}

In the first experiment, we obtain the transformation matrix $\mathbf{Q}$, using  CBOW and SVD embeddings separately, for the selected time periods,   $T_b=1930-1939$ and $T_t=1980-1989$.   \textbf{Table \ref{tab:1930-1980}} and \textbf{Table \ref{tab:1930-1980_SVD}} show the results for CBOW and SVD embeddings, respectively. 
We observe that OP+SC outperforms other methods in based on Recall@1, Recall@10, and MRR metrics. Regarding Recall@100 metric, OP outperforms OP+SC with SVD embeddings and achieves the same score with CBOW embeddings. 
LT  has the lowest scores in all cases, pointing out the  effectiveness of our proposed methods.

Regarding the impact of embeddings, we observe that when we use SVD embeddings Recall@1 and Recall@10  scores of OP and OP+SC  slightly decrease and their Recall@100 score slightly increases. However, the performance of LT  highly improves  in all metrics when SVD embeddings are used. 



\setlength\extrarowheight{5pt}
\begin{table}[!htb]
\normalsize
    \centering
    \begin{tabular}{p{3.8em}|p{3.8em}|p{3.8em}|p{3.8em}|p{3.8em}}
        \textbf{Method} & \textbf{Recall} @1 & \textbf{Recall} @10 & \textbf{Recall} @100 & \textbf{MRR} \\
        \hline
         LT & 0.09 & 0.34 & 0.61 & 0.17 \\
         OP &  0.33 &  0.71 & 0.84 & 0.45\\
         OP+SC &  \textbf{0.35} &  \textbf{0.74}& \textbf{0.84} & \textbf{0.81}\\
         \hline
    \end{tabular}
    \caption{Test results in finding counterparts of the words when the target period is $1980-1989$ and the base period is $1930-1939$ using CBOW embeddings. }
    \label{tab:1930-1980}
\end{table}

\setlength\extrarowheight{5pt}
\begin{table} [!htb]
\normalsize
    \centering
    \begin{tabular}{p{3.8em}|p{3.8em}|p{3.8em}|p{3.8em}|p{3.8em}}
        \textbf{Method} & \textbf{Recall} @1 & \textbf{Recall} @10 & \textbf{Recall} @100 & \textbf{MRR} \\
        \hline
         LT & 0.23 & 0.59 & 0.82 & 0.34 \\
         OP &  0.32 &  0.66 & \textbf{0.86} & 0.44\\
         OP+SC &  \textbf{0.33} &  \textbf{0.70}& 0.85 & \textbf{0.85}\\
         \hline
    \end{tabular}
    \caption{Test results in finding counterparts of the words when the target period is $1980-1989$ and the base period is $1930-1939$  using SVD embeddings. }
    \label{tab:1930-1980_SVD}
\end{table}



\begin{figure}[!htb]
    \centering
    \includegraphics[width=1.06\linewidth]{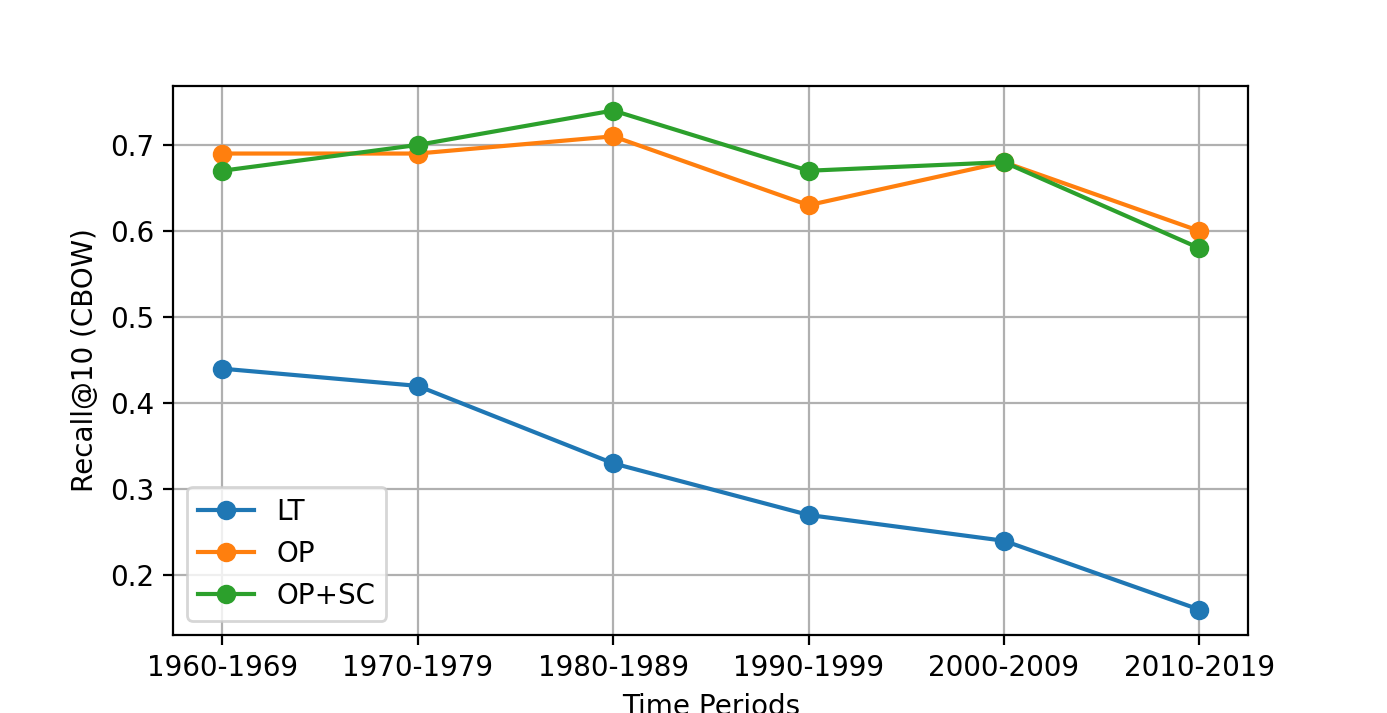}
    \caption{Performance of the methods across time periods in Recall@10 with CBOW embeddings.}
    \label{fig:timeseries-cbow}
\end{figure}

\begin{figure}[!htb]
    \centering
    \includegraphics[width=1\linewidth]{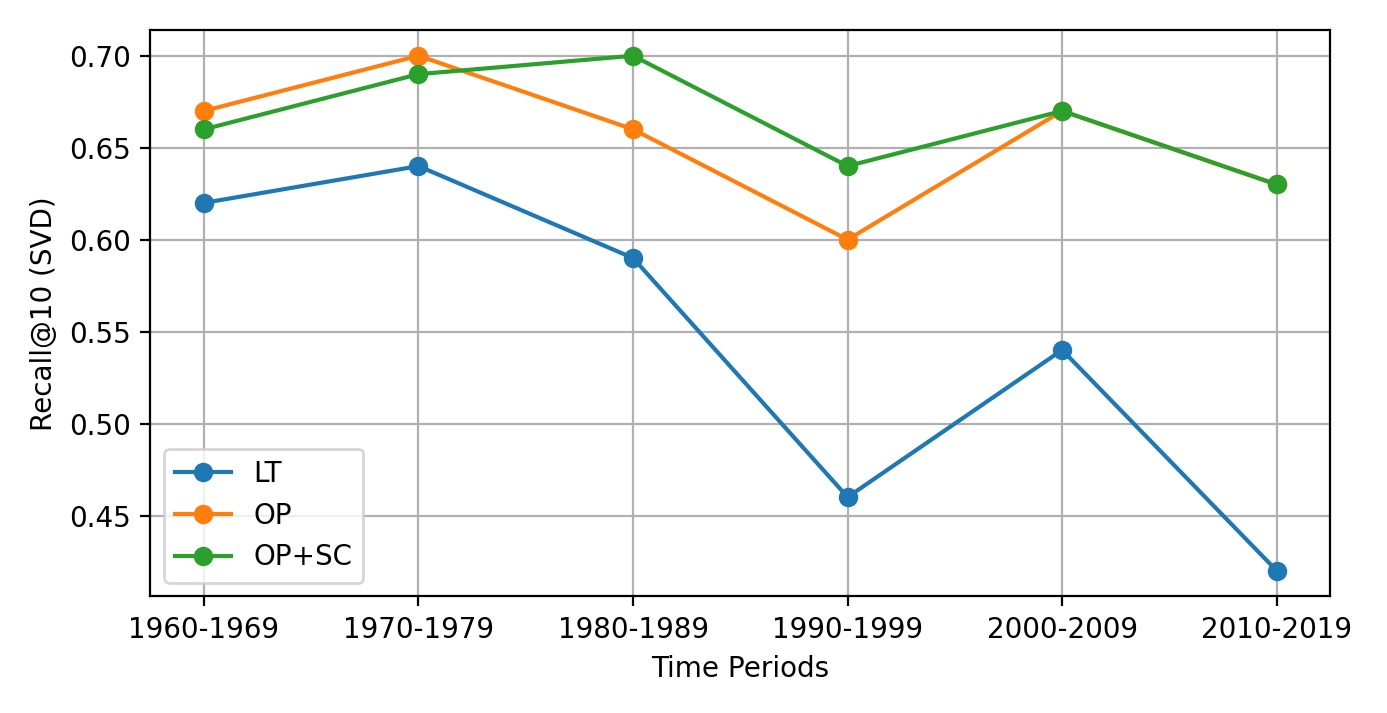}
    \caption{Performance of the methods across time periods in Recall@10 with SVD embeddings.}
    \label{fig:timeseries-svd}
\end{figure}

In our next experiment, we investigate how the temporal distance from the base time period to the target time period affects the results. In particular, we select the base period as $T_b=1930-1939$ and the target periods as $T_{t_1}=1960-1969$, $T_{t_2}=1970-1979$, $T_{t_3}=1980-1989$, $T_{t_4}=1990-1999$, $T_{t_5}=2000-2009$, and $T_{t_6}=2010-2019$. In the selection of the target periods, we ensured that the number of tokens was similar across the target periods to make the comparisons fair. Next, the pairs $(w,w')$ in the test set are filtered out by keeping only the pairs such that $w$ is present in all  target periods to make the results  directly comparable. Consequently, 220 pairs are left out of the original 221 pairs. We measure the recall@10 scores of LT, OP, OP+SC against different time periods. \textbf{Figure \ref{fig:timeseries-cbow}} and \textbf{Figure \ref{fig:timeseries-svd}} show the results for CBOW and SVD embeddings, respectively.

We have the following observations based on the results. Firstly, our results  suggest that OP and OP+SC are more robust against temporal differences when compared to LT. Secondly, we achieve similar results with both CBOW and SVD embeddings for our proposed methods. Lastly, the performance of LS tends to decrease as the time difference between the base period and the target period increases while the scores for our methods remains similar when the target time period is changed from 1960s to 1980s. However, their scores also slightly decreases for the subsequent time periods. 
This may be the result of the inherent characteristics of the language itself. In particular,  as the distance between the base period and the target period increases, the similarity between the vocabularies of the two time periods decreases. Some words from the previous periods disappear in the target period as time goes on. Consequently, the number of shared words between the two periods also decreases. This means that, based on Equation (\ref{eq:op}), the amount of data used to train the transformation matrices, $\mathbf{Q}$ and $\mathbf{M}$, is decreased. Furthermore, words undergo semantic changes over time. From the perspective of word embeddings, the phenomenon is explained by the shift of word vectors in the semantic space over time. This may also increase the error of the transformation matrix.

\section{Limitations}\label{sec:limitations}
In our experiments,  we  show that our methods are effective in extracting the old counterparts of modern Turkish words. However, our work can be improved in several directions. 
Firstly,  Turkronicles dataset 
contains many  noise words which seem to be due to the limitations of methods used to extract texts from documents during the creation of the dataset.  These words affect the generation of the similarity set of a given word. For instance, we noticed that the word \textit{vesik}, which is probably the misspelled version of \textit{vesika}, is found in the set of 10-most similar words for the word \textit{belge} 
when we used our OP method. Therefore, there might be many words which are actually correct but considered as false due to these noise words. Furthermore, noise words mislead the transformations in our methods. 
As another issue, we use unigram frequency to calculate  Spearman rank correlation. This method can be improved by considering  the co-occurrence with its context words.

Lastly, our experiments are based on a test set created using a single diachronic corpus. While Turkronicles is the largest and comprehensive diachronic corpus for Turkish, the  documents in the corpus do not reflect the whole characteristics of Turkish language and cover only the last 100 years. Therefore, our experiments have to be repeated on a larger dataset whenever such a diachronic corpus is created. Furthermore, extending the manually created test set will be beneficial to have a more reliable results.  

\section{Conclusion}\label{sec:conclusion}
The Turkish language has undergone several changes from the 1920s to the present day and it continues to evolve. 
Specifically, many new Turkish-originated words have been coined in the language, causing the Arabic-Persian origin words to be obsolete.  
In this study, we propose two methods based on orthogonal transformations to find the old counterparts of the modern Turkish words.  We conducted several experiments to compare our methods, assess how their performance is affected by the different embedding methods and difference between the time periods. 

Our analysis shows that orthogonal transformation is better than linear transformation in this task. Furthermore, sorting the similarity set of the aligned query word by their Spearman rank correlation values  improves the performance of the OP method in both SVD embeddings and CBOW embeddings in most of the cases.  

In the future, we plan to extend our work in several directions. We plan to increase our datasets and assess our methods in other languages. Furthermore, we plan to explore how our methods can be used to improve the performance of downstream tasks for  historical documents.



\begin{thebibliography}{1}

\bibitem{catastrophic}
Lewis, Geoffrey. The Turkish language reform: A catastrophic success: A catastrophic success. OUP Oxford, 1999.

\bibitem{lex}
Fischer. Roswitha, "Lexical change in present-day English: A corpus-based study of the motivation, institutionalization, and productivity of creative neologisms". Vol. 17.  Gunter Narr Verlag, 1998.

\bibitem{neo_def}
Rodríguez Guerra, Alexandre. "Dictionaries of Neologisms: a Review and Proposals for its Improvement" Open Linguistics, vol. 2, no. 1, 2016. https://doi.org/10.1515/opli-2016-0028

\bibitem{neo_1}
John R Taylor and Anthony P. Grant. 2014. Lexical Borrowing. Oxford University Press, Oxford.

\bibitem{pechenick}
Pechenick, Eitan Adam, Christopher M. Danforth, and Peter Sheridan Dodds. "Characterizing the Google Books corpus: Strong limits to inferences of socio-cultural and linguistic evolution." PloS one 10.10 (2015): e0137041.

\bibitem{michel}
Michel, Jean-Baptiste, et al. "Quantitative analysis of culture using millions of digitized books." science 331.6014 (2011): 176-182.

\bibitem{tredici}
Del Tredici, Marco, and Raquel Fernández. "The road to success: Assessing the fate of linguistic innovations in online communities." arXiv preprint arXiv:1806.05838 (2018).

\bibitem{progress}
Aitchison. J, "The reason why", \emph{Language change: Progress or decay?}, ISBN: 9781107023628,  Cambridge University Press, 2012.

\bibitem{kulkarni}
Kulkarni, Vivek, et al. "Statistically significant detection of linguistic change." Proceedings of the 24th international conference on world wide web. 2015.

\bibitem{hierarchy}
Zhang, Yating, Adam Jatowt, and Katsumi Tanaka. "Temporal analog retrieval using transformation over dual hierarchical structures." Proceedings of the 2017 ACM on Conference on Information and Knowledge Management. 2017.

\bibitem{levy}
Levy, Omer, Yoav Goldberg, and Ido Dagan. "Improving distributional similarity with lessons learned from word embeddings." Transactions of the association for computational linguistics 3 (2015): 211-225.

\bibitem{kutuzov}
Kutuzov, Andrey, et al. "Diachronic word embeddings and semantic shifts: a survey." arXiv preprint arXiv:1806.03537 (2018).

\bibitem{hamilton_a}
Hamilton, William L., Jure Leskovec, and Dan Jurafsky. "Diachronic word embeddings reveal statistical laws of semantic change." arXiv preprint arXiv:1605.09096 (2016).

\bibitem{turkronicles}
Yazar,  Togay,  Mucahid Kutlu, and İsa Kerem Bayırlı. "Turkronicles: Diachronic Resources for the Fast Evolving Turkish Language." arXiv preprint arXiv:2405.10133 (2024).

\bibitem{hamilton_b}
Hamilton, William L., Jure Leskovec, and Dan Jurafsky. "Diachronic word embeddings reveal statistical laws of semantic change." arXiv preprint arXiv:1605.09096 (2016)

\bibitem{bamman}
Bamman, David, and Gregory Crane. "Measuring historical word sense variation." Proceedings of the 11th annual international ACM/IEEE joint conference on Digital libraries. 2011.

\bibitem{aitc}
Denison, David. "Jean Aitchison, Language change: progress or decay?(Fontana Linguistics.) London: Fontana, 1981. Pp. 266." Journal of Linguistics 19.2 (1983): 503-504.

\bibitem{goel}
Goel, Anmol, and Ponnurangam Kumaraguru. "Detecting Lexical Semantic Change across Corpora with Smooth Manifolds (Student Abstract)." Proceedings of the AAAI Conference on Artificial Intelligence. Vol. 35. No. 18., 2021.
\bibitem{bybee}
Bybee, Joan. Language change. Cambridge University Press, 2015.

\bibitem{mikolov}
Mikolov, Tomas, et al. "Efficient estimation of word representations in vector space." arXiv preprint arXiv:1301.3781 (2013).

\bibitem{xu}
Xu, Yang, and Charles Kemp. "A Computational Evaluation of Two Laws of Semantic Change." CogSci. 2015.

\bibitem{schonemann}
Schönemann, Peter H. "A generalized solution of the orthogonal procrustes problem." Psychometrika 31.1 (1966): 1-10.

\bibitem{rudolph}
Rudolph, Maja, and David Blei. "Dynamic embeddings for language evolution." Proceedings of the 2018 world wide web conference. 2018.

\bibitem{neology}
Ryskina, Maria, et al. "Where new words are born: Distributional semantic analysis of neologisms and their semantic neighborhoods." arXiv preprint arXiv:2001.07740 (2020).

\bibitem{omni}
Zhang, Yating, et al. "Omnia mutantur, nihil interit: Connecting past with present by finding corresponding terms across time." Proceedings of the 53rd Annual Meeting of the Association for Computational Linguistics and the 7th International Joint Conference on Natural Language Processing (Volume 1: Long Papers). 2015.

\bibitem{entity}
Onoe, Yasumasa, et al. "Entity cloze by date: What LMs know about unseen entities." arXiv preprint arXiv:2205.02832 (2022).

\bibitem{dyn1}
Yao, Zijun, et al. "Dynamic word embeddings for evolving semantic discovery." Proceedings of the eleventh acm international conference on web search and data mining. 2018.
\bibitem{dyn2}
Di Carlo, Valerio, Federico Bianchi, and Matteo Palmonari. "Training temporal word embeddings with a compass." Proceedings of the AAAI conference on artificial intelligence. Vol. 33. No. 01. 2019.

\bibitem{temporal_mis}
Luu, Kelvin, et al. "Time waits for no one! analysis and challenges of temporal misalignment." arXiv preprint arXiv:2111.07408 (2021).

\bibitem{test_data}
Zheng, Jonathan, Alan Ritter, and Wei Xu. "NEO-BENCH: Evaluating Robustness of Large Language Models with Neologisms." arXiv preprint arXiv:2402.12261 (2024).



\end{thebibliography}
\end{document}